%% file: main.tex
\documentclass[10pt,twocolumn,letterpaper]{article}
\usepackage{multirow}
\usepackage[pagenumbers]{cvpr} 

\input{preamble}

\definecolor{cvprblue}{rgb}{0.21,0.49,0.74}
\usepackage[pagebackref,breaklinks,colorlinks,allcolors=cvprblue]{hyperref}

\def\paperID{1916} 
\def\confName{CVPR}
\def\confYear{2026}

\usepackage{dsfont}
\title{FCL-COD: Weakly Supervised Camouflaged Object Detection with Frequency-aware and Contrastive Learning}

\author{
    \textbf{Jingchen Ni}$^{1*}$ \quad
    \textbf{Quan Zhang}$^{1*}$ \quad
    \textbf{Dan Jiang}$^{1}$ \quad
    \textbf{Keyu Lv}$^{1}$ \quad
    \textbf{Ke Zhang}$^{2\dagger}$ \quad
    \textbf{Chun Yuan}$^{1\dagger}$
    \\[.6em]
    $^{1}$Tsinghua University \quad
    $^{2}$Soochow University 
    \\[.5em]
    \tt\small
    njc24@mails.tsinghua.edu.cn, zhangqua22@tsinghua.org.cn,\\
    \tt\small
    kzhang19@suda.edu.cn, yuanc@sz.tsinghua.edu.cn
}

\begin{document}
\maketitle
\begingroup
\renewcommand{\thefootnote}{\fnsymbol{footnote}}
\footnotetext[1]{Equal contribution; $^{\dagger}$ corresponding authors.}
\endgroup
\input{sec/0_abstract}    
\input{sec/1_intro}
\input{sec/2_related_work}
\input{sec/3_method}
\input{sec/4_experiment}
\input{sec/5_conclusion}

{
    \small
    \bibliographystyle{ieeenat_fullname}
    \bibliography{main}
}


\end{document}

%% file: preamble.tex
\usepackage{pifont}
\usepackage{float}



%% file: sec/0_abstract.tex
\begin{abstract}
Existing camouflage object detection (COD) methods typically rely on fully-supervised learning guided by mask annotations. However, obtaining mask annotations is time-consuming and labor-intensive. Compared to fully-supervised methods, existing weakly-supervised COD methods exhibit significantly poorer performance. Even for the Segment Anything Model (SAM), there are still challenges in handling weakly-supervised camouflage object detection (WSCOD), such as: a. non-camouflage target responses, b. local responses, c. extreme responses, and d. lack of refined boundary awareness, which leads to unsatisfactory results in camouflage scenes.  To alleviate these issues, we propose a frequency-aware and contrastive learning-based WSCOD framework in this paper, named FCL-COD. To mitigate the problem of non-camouflaged object responses, we propose the Frequency-aware Low-rank Adaptation (FoRA) method, which incorporates frequency-aware camouflage scene knowledge into SAM. To overcome the challenges of local and extreme responses, we introduce a gradient-aware contrastive learning approach that effectively delineates precise foreground-background boundaries. Additionally, to address the lack of refined boundary perception, we present a multi-scale frequency-aware representation learning strategy that facilitates the modeling of more refined boundaries. We validate the effectiveness of our approach through extensive empirical experiments on three widely recognized COD benchmarks. The results confirm that our method surpasses both state-of-the-art weakly supervised and even fully supervised techniques.
\end{abstract}

%% file: sec/1_intro.tex
\section{Introduction}
\label{sec:intro}

\begin{figure}[t]
    \centering
    \includegraphics[width=1.0\linewidth]{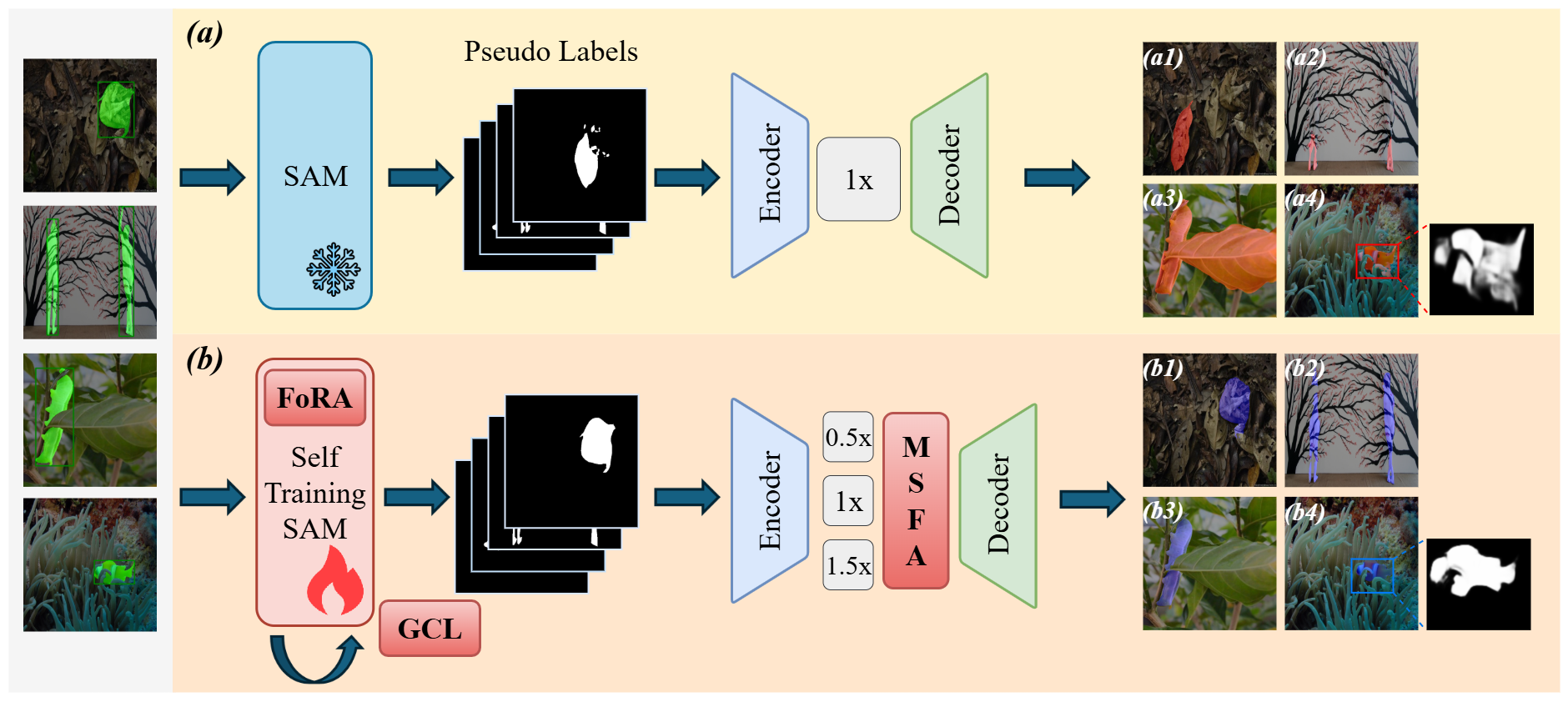}
    \caption{Comparison with the baseline,where (a) previous SAM-based method and (b) shows our proposed FCL-COD. By incorporating frequency awareness and contrastive learning, FCL-COD avoids the baseline’s drawbacks of a1) responses to non-camouflaged objects, a2) localized responses, a3) extreme responses, and a4) coarse boundaries.}
    \label{fig:introduction}
    \end{figure}
Camouflaged Object Detection (COD) aims to identify and segment objects concealed within their surrounding environment, attracting significant attention due to its potential applications in fields such as medical diagnosis\cite{fan2020inf}, species conservation\cite{ji2022video}, and crop pest detection\cite{nafus2015hiding}. Unlike traditional object detection\cite{fan2022salient,he2017mask}, COD faces the challenge of high intrinsic similarity between the camouflaged object and its background, requiring the recognition of internal object information based on fine-grained details. Additionally, as a pixel-level classification task, COD demands precise boundary detection results.

In recent years, there has been an increasing amount of research on COD using data-driven deep learning techniques. Although fully supervised methods that rely on pixel-level annotations have made progress, they face inherent obstacles: manual pixel-level annotation of large-scale datasets is time-consuming and labor-intensive, and traditional methods treat each pixel in the target area equally, potentially failing to capture the essential structural features of the object\cite{liu2019pestnet}. To overcome these obstacles, sparse annotation methods have emerged to simplify dataset annotation and reduce overfitting. Exploring the use of sparse annotations as supervision in Weakly Supervised Camouflaged Object Detection (WSCOD) has become a promising approach. Recent progress in weakly supervised, zero-shot, and multimodal localization under sparse supervision~\cite{zhang2025weakly,11210145,wang2025iterprime} further highlights the potential of learning reliable representations from limited annotations.

However, the limited annotation information available in WSCOD can severely hinder detection performance. While weakly supervised methods offer flexibility and reduce dataset annotation time and labor costs, they often fail to provide sufficient cues for accurate boundary inference. To address this issue, some studies\cite{he2023weakly} introduce consistency constraints, yet these losses are still computed on weakly represented image features and may therefore lead to imprecise object localization. Although boundary priors can partially improve contour quality, the cluttered backgrounds in camouflage scenes often introduce substantial non-target noise. As a result, existing WSCOD methods still suffer from non-camouflaged object responses, local responses, extreme responses, and insufficient boundary perception. Such challenges are also prevalent in broader weakly supervised and cross-modal settings~\cite{zhang2025rethinking,11084533,yan2026learning,ni2025semantic}, motivating the design of more robust frequency-aware representations.

As a universal segmentation foundation model, SAM~\cite{kirillov2023segment} has shown strong adaptability across downstream dense prediction tasks such as image restoration~\cite{zhang2024distilling}, image manipulation detection~\cite{zhang2025imdprompter}, and segmentation~\cite{wang2025sam2,wang2024convolution}. Related advances in image manipulation detection and localization~\cite{chen2025gim} further highlight the promise of foundation-model-based fine-grained visual understanding, making SAM a compelling backbone for weakly supervised scenarios. To address the issues present in previous Weakly Supervised Camouflaged Object Detection (WSCOD) methods, we propose a frequency-aware and contrastive learning-based weakly supervised camouflaged object detection framework, FCL-COD. To tackle the problem of non-camouflaged object responses, we introduce frequency-aware low-rank adaptation to inject camouflaged object scene knowledge into the pre-trained backbone model SAM. For the issues of local responses and extreme responses, we propose gradient-aware contrastive learning, which explores difficult background areas through a gradient-aware strategy and increases the representation distance between the foreground and background in the high-dimensional space. To address the lack of refined boundary perception, we propose a multi-scale frequency-aware attention module that combines multi-scale attention perception in both the frequency and spatial domains to uncover boundary-sensitive feature representations.

In summary, our contributions are as follows:
\begin{itemize}
    \item We propose a frequency-aware and contrastive learning-based WSCOD method, which explores fine-grained object boundaries by mining high-dimensional frequency-domain differences, and uses contrastive learning to separate the object and background in the representation space.
    \item We introduce frequency-aware low-rank adaptation, which injects frequency-aware camouflaged object knowledge into SAM. Combined with gradient-aware contrastive learning, we mine easily confused background areas to push the object and background apart in the high-dimensional representation space.
    \item We propose a multi-scale frequency-aware attention module, which realizes boundary-sensitive representation learning through multi-scale interactions between the frequency and spatial domains.
    \item Extensive empirical experiments were conducted on four mainstream COD benchmarks. The results show that FCL-COD outperforms the state-of-the-art weakly supervised methods, and even fully supervised methods.
\end{itemize}

%% file: sec/2_related_work.tex
\section{Related Work}
\label{sec:related_work}

\subsection{Camouflaged Object Detection}

Traditional camouflaged object detection (COD) methods rely on handcrafted features such as color \cite{huerta2007improving}, texture \cite{galun2003texture}, and other visual cues. However, their performance degrades in complex scenes where camouflaged objects closely resemble the background. With the rise of deep learning\cite{qiu2024tfb,qiu2025duet,yan2026adamem,qiu2025DBLoss,qiu2025dag,qiu2025tab} and the establishment of benchmark datasets \cite{fan2020camouflaged}, deep learning-based COD approaches have gained prominence. Some methods \cite{fan2020camouflaged} draw inspiration from biological perception, such as SINet, which mimics predators’ two-stage hunting strategy. Others employ multi-task frameworks that jointly learn COD and edge detection, using graphical models to capture task dependencies \cite{zhai2021mutual}. To overcome the limitations of RGB-only input, recent studies \cite{wang2023depth, zhong2022detecting} integrate auxiliary cues like frequency-domain features to better distinguish subtle foreground–background differences. Despite the advances of fully supervised models, their dependence on dense pixel-level annotations often overlooks holistic target structures, motivating the development of weakly supervised COD methods that exploit sparse supervision for more efficient learning.

\subsection{Contrastive Learning}

\begin{figure*}[t]
\centering
\includegraphics[height=7cm]{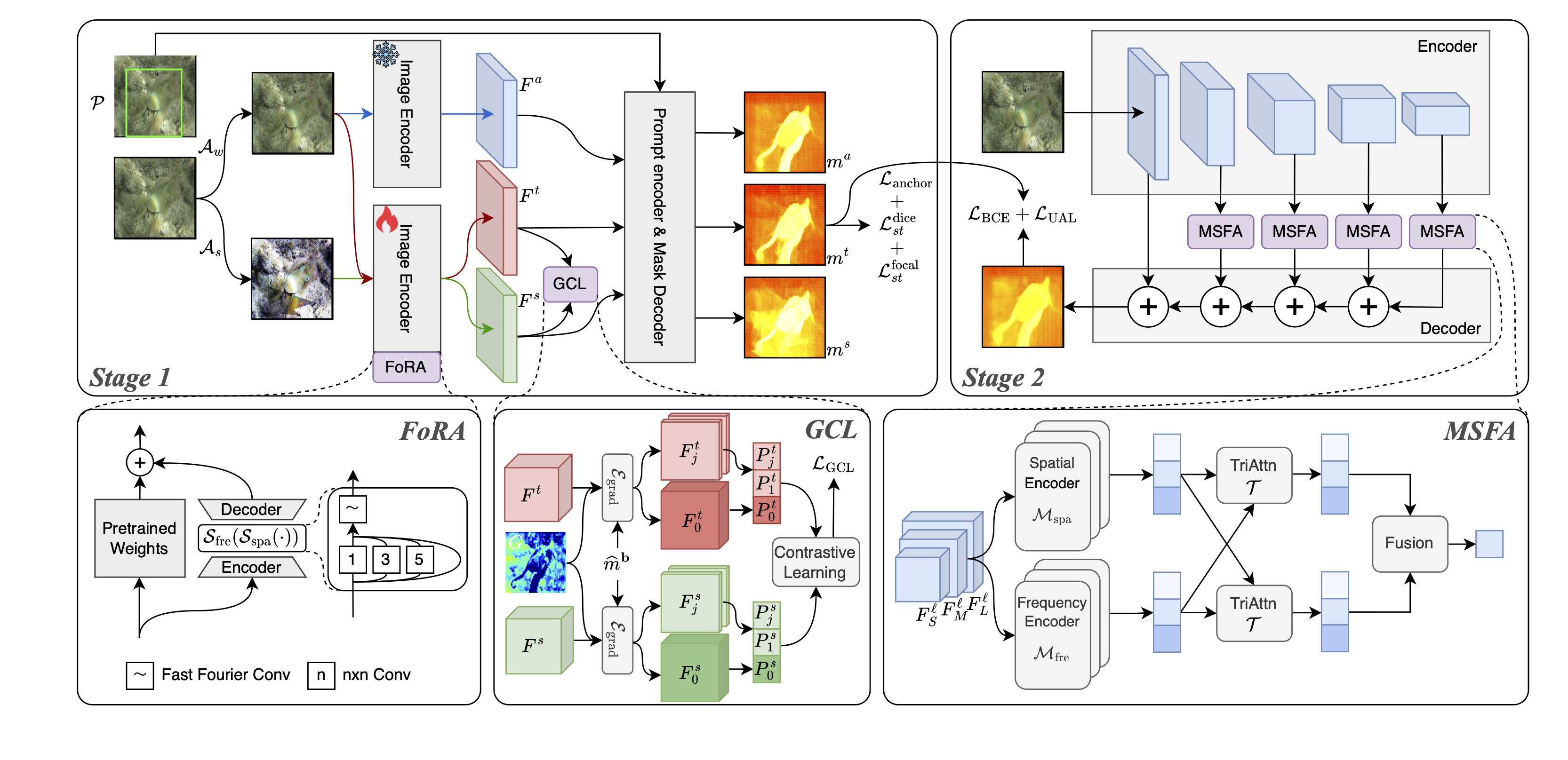}
\caption{\textbf{FCL-COD Pipeline Overview.} Within a triadic teacher–student architecture, frequency-aware low-rank adaptation and gradient-aware contrastive learning jointly suppress the localized and extreme activations that plague earlier approaches. In the subsequent re-training phase, multi-scale frequency-aware attention is injected to excavate crisper, more delicate camouflaged boundaries.}
\label{fig:pipeline}
\end{figure*}

Recently, contrastive learning has become a dominant self-supervised paradigm for learning discriminative representations from unlabeled data \cite{caron2020unsupervised}. By pulling similar samples together and pushing dissimilar ones apart, it effectively enhances feature separability—an essential property for camouflaged object detection. Representative methods include SimCLR \cite{chen2020simple}, which employs data augmentation and transformation prediction to learn invariant representations; MoCo \cite{he2020momentum}, which introduces a momentum-updated encoder and dynamic queue to improve negative sampling; and SwAV \cite{caron2020unsupervised}, which integrates clustering with contrastive objectives for fine-grained semantic learning. Moreover, InfoNCE \cite{rusak2024infonce} formalizes contrastive learning via mutual information maximization, reinforcing affinity preservation in unsupervised representation learning. These advances underscore contrastive learning’s strength in boosting generalization and feature discriminability. Nevertheless, its application to COD remains challenging due to the difficulty of defining positive/negative pairs and efficiently mining hard negatives—issues that merit further exploration.

%% file: sec/3_method.tex
\section{Method}
\label{sec:method}

Our proposed \textbf{FCL-COD} employs a two-stage framework. In the first stage, SAM is adapted to produce high-quality pseudo-labels for camouflaged objects via \textit{Triadic Teacher–Student Self-training} (\S\ref{subsec:triadic}), enhanced by \textit{Frequency-aware Low-Rank Adaptation} (FoRA, \S\ref{subsec:fora}) and \textit{Gradient-aware Contrastive Learning} (GCL, \S\ref{subsec:gcl}). In the second stage, a lightweight encoder–decoder detector with \textit{Multi-Scale Frequency-aware Attention} (MSFA, \S\ref{subsec:msfa}) is trained using these pseudo-labels. This design achieves an optimal trade-off between accuracy and efficiency: SAM offers robust priors for pseudo-labeling, while the MSFA-equipped lightweight detector enables real-time inference with boundary-sensitive features.

\subsection{Triadic Teacher-Student Self-training}
\label{subsec:triadic}
To obtain reliable pseudo-labels, three encoders are maintained: an anchor encoder $f^a(x;\Theta^a)$, a student encoder $f^s(x;\Theta^s)$, and a teacher encoder $f^t(x;\Theta^t)$, where $\Theta^s = \Theta^t$. As shown in \cref{fig:pipeline}, for each sample $x_i$, weak augmentation $\mathcal{A}_w$ is applied to anchor and teacher inputs, and strong augmentation $\mathcal{A}_s$ to the student input. The encoders yield feature maps $F^a$, $F^s$, and $F^t \in \mathbb{R}^{D \times H \times W}$. Given a bounding box prompt $\mathcal{P}$, each branch produces $N_p$ masks ${y_j^a}$, ${y_j^s}$, ${y_j^t}$, $j=1,\dots,N_p$, which are normalized via sigmoid to $m_j \in [0,1]^{H \times W}$ and binarized as $\widehat{m}_j = \mathds{1}(m_j > 0.5) \in {0,1}^{H \times W}$.

The triadic self-training loss combines two terms. The \textit{student–teacher loss} employs Focal Loss and Dice Loss to guide the student with the teacher’s pseudo-labels, where $\gamma$ emphasizes hard pixels and $\epsilon$ prevents division by zero. The Focal Loss is formulated in \cref{eq:st_focal}.

\begin{equation}
\small
\begin{aligned}
\mathcal{L}_{st}^{\text{focal}} &= -\frac{1}{HW} \sum_{j=1}^{N_p} \sum_{h,w} \mathds{1}(\hat{m}_{jhw}^{t}=1) \bigl( 1 - m_{jhw}^{s} \bigr)^{\gamma} \log \bigl( m_{jhw}^{s} \bigr) \\
& \quad + \mathds{1}(\hat{m}_{jhw}^{t}=0) \bigl( m_{jhw}^{s} \bigr)^{\gamma} \log \bigl( 1 - m_{jhw}^{s} \bigr)
\end{aligned}
\label{eq:st_focal}
\end{equation}

The Dice Loss is given by Eq.~(\ref{eq:st_dice}).
\begin{equation}
    \mathcal{L}_{st}^{dice}=\sum_{j=1}^{N_p} 1-\frac{2 \sum_{h, w} m_{j h w}^s \cdot \widehat{m}_{j h w}^t+\epsilon}{\sum_{h, w} m_{j h w}^s+\sum_{h, w} \widehat{m}_{j h w}^t+\epsilon}
    \label{eq:st_dice}
\end{equation}

Second, to mitigate error accumulation caused by imperfect teacher pseudo-labels, we introduce an \textit{anchor loss} for regularization. A frozen anchor network preserves the original SAM knowledge and constrains both the student and teacher, preventing excessive deviation from the pre-trained model. The anchor loss is formulated in Eq.~(\ref{eq:anchor}).

\begin{equation}
\mathcal{L}_{\text{anchor}}=\lambda_{\text{stu}}^{\text{dice}} \mathcal{L}^{\text{dice}}\left(m^s, \widehat{m}^a\right)+\lambda_{\text{tea}}^{\text{dice}} \mathcal{L}^{\text{dice}}\left(m^t, \widehat{m}^a\right)
\label{eq:anchor}
\end{equation}

\subsection{Frequency-aware Low-Rank Adaptation}
\label{subsec:fora}

Low-Rank Adaptation (LoRA) constrains parameter updates to a low-dimensional subspace via an encoder–decoder parameterization. While freezing the pre-trained parameters, it injects a pair of lightweight, trainable rank-decomposition matrices into each Transformer layer. Given a linear projection weight $W_0 \in \mathbb{R}^{b \times a}$, LoRA introduces two trainable matrices $W_e \in \mathbb{R}^{r \times a}$ and $W_d \in \mathbb{R}^{b \times r}$, where $r \ll \min(a, b)$. The forward propagation is given by Eq.~(\ref{eq:lora_forward}).
\begin{equation}
    h = W_0 x + W_d W_e x
    \label{eq:lora_forward}
\end{equation}

The low-rank branch $W_d W_e x$ serves as a residual update to the frozen projection $W_0 x$.

However, in camouflaged-object scenarios, where boundaries are ambiguous and transparent targets often overlap with cluttered backgrounds, conventional low-rank adaptation struggles to inject camouflage-specific priors into SAM, leading to weak discrimination and spurious activations in non-camouflaged regions. To address this, we propose Frequency-aware Low-Rank Adaptation (FoRA), which extends LoRA by inserting a two-stage transformation between the encoder and decoder to enrich the encoded features in both spatial and frequency domains.

Formally, the FoRA-enhanced forward propagation is defined in Eq.~(\ref{eq:fora_forward}).
\begin{equation}
    h = W_0 x + W_d \, \mathcal{S}_{\text{fre}}\left(\mathcal{S}_{\text{spa}}(W_e x)\right)
    \label{eq:fora_forward}
\end{equation}
The spatial operator precedes the frequency operator within the low-rank pathway.
Here, \( \mathcal{S}_{\text{spa}}(\cdot) \) denotes the spatial enhancement stage and \( \mathcal{S}_{\text{fre}}(\cdot) \) denotes the frequency modulation stage. The spatial enhancement stage captures multi-scale contextual dependencies via convolutions with different receptive fields. Given the encoded feature $F_e = W_e x$, we define the spatial enhancement in Eq.~(\ref{eq:spatial}), which aggregates multi-scale responses and adds a residual connection.
\begin{equation}
    \mathcal{S}_{\text{spa}}(F_e) = \frac{\Phi_{1\times1}(F_e) + \Phi_{3\times3}(F_e) + \Phi_{5\times5}(F_e)}{3} + F_e
    \label{eq:spatial}
\end{equation}

Subsequently, the frequency modulation stage applies the Fourier transform, performs convolution in the frequency domain, and reconstructs the representation via the inverse Fourier transform. The transformation is given in Eq.~(\ref{eq:frequency}).
\begin{equation}
    \mathcal{S}_{\text{fre}}(F) = \text{IFFT}\left(\Phi_{3\times3}\left(\text{FFT}(F)\right)\right)
    \label{eq:frequency}
\end{equation}

By explicitly modeling spatial and frequency cues in a cascaded manner as shown in Eq.~(\ref{eq:fora_forward}), FoRA augments the base projection with spatial–frequency enriched low-rank updates and more effectively injects camouflage-specific priors while preserving the generalization ability of the foundation model. 

\subsection{Gradient-aware Contrastive Learning}
\label{subsec:gcl}

Camouflaged scenes often contain ambiguous boundaries between foreground and background, which can lead to partial detections and a high rate of false positives. While the losses in the triadic teacher–student framework operate at the output level, they are insufficient to enforce separability in the feature space. To further enhance the distinction between foreground and background representations, we introduce a Gradient-aware Contrastive Learning (GCL) objective.

During contrastive learning, the quality of sampled features plays a crucial role. To emphasize ambiguous background regions that are easily confused with the foreground, we derive a gradient activation map $G^t$ via Grad-CAM from the teacher feature map $F^t$, which provides more stable and reliable guidance than the student. The gradient activation map is computed as shown in Eq.~(\ref{eq:gcl_grad}), and we construct a gradient-weighted background mask according to Eq.~(\ref{eq:gcl_bgmask}):
\begin{equation}
    G^t = \Phi_{\text{GC}}(F^t), \quad G^t \in [0,1]^{H \times W}
    \label{eq:gcl_grad}
\end{equation}
\begin{equation}
    \widetilde{m}_0 = \widehat{m}_0 \odot G^t
    \label{eq:gcl_bgmask}
\end{equation}
where $\odot$ denotes element-wise multiplication, $\widehat{m}_0$ is the binary background mask, and $\widetilde{m}_0$ is its gradient-weighted counterpart.

We then compute branch-specific instance prototypes via masked average pooling. For each branch $\mathbf{b} \in \{s, t\}$, the background prototype and the $j$-th foreground prototype are defined as. Here, the superscripts $s$ and $t$ indicate the student and teacher branches, respectively; the corresponding feature maps are $F^s$ and $F^t$, and the instance prototypes are $I_j^s$ and $I_j^t$.
\begin{equation}
    I_0^{\mathbf{b}} = \frac{\sum_{h, w} F^{\mathbf{b}}_{h w} \cdot \widetilde{m}_{0 h w}}{\sum_{h, w} \widetilde{m}_{0 h w}}, \quad \mathbf{b} \in \{s,t\}
    \label{eq:gcl_bgpool}
\end{equation}
\begin{equation}
    I_j^{\mathbf{b}} = \frac{\sum_{h, w} F^{\mathbf{b}}_{h w} \cdot \widehat{m}_{j h w}}{\sum_{h, w} \widehat{m}_{j h w}}, \quad \mathbf{b} \in \{s,t\},\ j=1,\dots, N
    \label{eq:gcl_fgpool}
\end{equation}
Here, $N$ denotes the number of foreground instances in the current image (variable across samples), and index $0$ refers to the background prototype.

Here, $F^{\mathbf{b}}$ denotes the L2-normalized feature map, i.e., $F^{\mathbf{b}}_{h w} = F^{\mathbf{b}}_{h w}/\lVert F^{\mathbf{b}}_{h w}\rVert_2$. In Eq.~(\ref{eq:gcl}), positive pairs are constructed between $I_j^s$ and $I_j^t$, while negatives include other foreground instances as well as the gradient-aware background prototype $I_0^t$. Weighting the background with $\widetilde{m}_0$ (Eqs.~(\ref{eq:gcl_bgmask})–(\ref{eq:gcl_bgpool})) encourages the model to focus on Grad-CAM–highlighted background regions that are most prone to confusion with the foreground. The contrastive objective is formulated as:
\begin{equation}
    \mathcal{L}_{\text{GCL}} = -\log \frac{\sum_{j=1}^{N} \exp \left(I_j^s \cdot I_j^t / \tau \right)}{\sum_{j=1}^{N} \sum_{\substack{k=0 \\ k \neq j}}^{N} \exp \left(I_j^s \cdot I_k^t / \tau \right)}
    \label{eq:gcl}
\end{equation}
where $\tau$ is the temperature hyperparameter.

Finally, the overall training objective integrates the gradient-aware contrastive loss with the teacher–student self-training and anchor losses, as defined in Eq.~(\ref{eq:gcl_final}).
\begin{equation}
    \mathcal{L} = \mathcal{L}_{\text{st}}^{\text{dice}} + \lambda_1 \mathcal{L}_{\text{anchor}} + \lambda_2 \mathcal{L}_{\text{GCL}} + \lambda_3 \mathcal{L}_{\text{st}}^{\text{focal}}
    \label{eq:gcl_final}
\end{equation}

By incorporating the gradient-aware contrastive term, the model achieves better separation between foreground and background features in the embedding space, leading to more robust discrimination of camouflaged targets.

\begin{figure}
    \centering
    \includegraphics[width=1\linewidth]{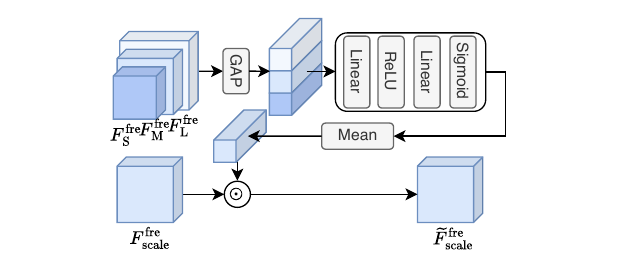}
    \caption{Detailed structure of Tri-Channel Attention mechanism in MSFA}
    \label{fig:subflow}
\end{figure}

\subsection{Multi-Scale Frequency-aware Attention}
\label{subsec:msfa}
The high-quality pseudo-labels generated in the first stage enable efficient training of a lightweight detector. However, camouflaged objects exhibit ambiguous boundaries that require fine-grained boundary-sensitive features. To address this, we propose MSFA as the second-stage module inserted between the encoder and decoder. For each encoder layer $\ell$, we extract multi-scale features at three different resolutions: small-scale $F_S^{\ell}$, medium-scale $F_M^{\ell}$, and large-scale $F_L^{\ell}$, which capture fine details, intermediate structures, and global context, respectively. These three-scale features are processed by MSFA before decoding. 

MSFA adopts a dual-branch design: the spatial branch $\mathcal{M}_{\text{spa}}$ enhances local context via stacked $3\times3$ convolutions, while the frequency branch $\mathcal{M}_{\text{fre}}$ models spectral cues in the Fourier domain, as defined in Eq.~(\ref{eq:msfa_branches}).
\begin{equation}
\begin{aligned}
    \mathcal{M}_{\text{spa}}(F) &= \Phi_{3\times3}(\Phi_{3\times3}(F)) \\
    \mathcal{M}_{\text{fre}}(F) &= \text{IFFT}(\Phi_{1\times1}([\Re(\text{FFT}(F)),\,\Im(\text{FFT}(F))]))
\end{aligned}
\label{eq:msfa_branches}
\end{equation}
Here, $\Phi_{k\times k}$ denotes a $k\times k$ convolution, and $\Re(\cdot)$, $\Im(\cdot)$ denote the real and imaginary parts of the FFT output, respectively.

To enable cross-domain interaction, we employ a Tri-Channel Attention mechanism, denoted as $\mathcal{T}$, to gate each branch with multi-scale context from the other domain. For each scale, we first apply the two branches to obtain $F_{\text{scale}}^{\text{spa}}=\mathcal{M}_{\text{spa}}(F_{\text{scale}}^{\ell})$ and $F_{\text{scale}}^{\text{fre}}=\mathcal{M}_{\text{fre}}(F_{\text{scale}}^{\ell})$. The gating operator $\mathcal{T}$ is defined in Eq.~(\ref{eq:msfa_attn}).
\begin{equation}
\small
    \mathcal{T}_i(x\,|\,\{y_1, y_2, y_3\}) = x \odot \tfrac{1}{3}\sum_{j=1}^{3}\sigma(W_{i,j}^{(2)} \cdot \text{ReLU}(W_{i,j}^{(1)} \cdot \text{GAP}(y_j)))
\label{eq:msfa_attn}
\end{equation}
where $\text{GAP}(\cdot)$ denotes global average pooling, $W_{i,j}^{(1)}$ and $W_{i,j}^{(2)}$ are scale-specific learnable weight matrices for channel reduction and expansion, $\sigma$ is the sigmoid activation, and the summation averages attention weights from three scales. Applying Eq.~(\ref{eq:msfa_attn}), we obtain the gated features as shown in Eq.~(\ref{eq:msfa_gate}).
\begin{equation}
\begin{aligned}
    \widetilde{F}_{\text{scale}}^{\text{fre}} &= \mathcal{T}_{\text{scale}}(F_{\text{scale}}^{\text{fre}}\,|\,\{F_S^{\text{spa}},F_M^{\text{spa}},F_L^{\text{spa}}\}) \\
    \widetilde{F}_{\text{scale}}^{\text{spa}} &= \mathcal{T}_{\text{scale}}(F_{\text{scale}}^{\text{spa}}\,|\,\{F_S^{\text{fre}},F_M^{\text{fre}},F_L^{\text{fre}}\})
\end{aligned}
\label{eq:msfa_gate}
\end{equation}
Here, the first line gates frequency features with spatial context, while the second line gates spatial features with frequency context, for each scale $\in\{S,M,L\}$. Finally, the gated features from both branches are aggregated and fused as shown in Eq.~(\ref{eq:msfa_overall}).
\begin{equation}
    F_{\text{MSFA}}^{\ell} = \Phi_{1\times1}^{\text{spa}}([\widetilde{F}_S^{\text{spa}},\widetilde{F}_M^{\text{spa}},\widetilde{F}_L^{\text{spa}}]) + \Phi_{1\times1}^{\text{fre}}([\widetilde{F}_S^{\text{fre}},\widetilde{F}_M^{\text{fre}},\widetilde{F}_L^{\text{fre}}])
\label{eq:msfa_overall}
\end{equation}
where $[\cdot]$ denotes channel-wise concatenation, and $\Phi_{1\times1}^{\text{spa}}$ and $\Phi_{1\times1}^{\text{fre}}$ are branch-specific projection convolutions. This design enables the model to capture boundary-sensitive representations through multi-scale spatial–frequency interaction.

\textbf{Training Objective.} In the second stage, we train the MSFA-enhanced detector using the pseudo-labels generated from the first stage. Given the predicted probability map $p$ and the pseudo-label mask $\widehat{m}$, the training objective combines binary cross-entropy loss and uncertainty-aware loss, defined in Eq.~(\ref{eq:stage2_loss}).
\begin{equation}
\begin{aligned}
    &\mathcal{L}_{\text{stage2}} = \mathcal{L}_{\text{BCE}}(p, \widehat{m}) + \alpha(t) \cdot \mathcal{L}_{\text{UAL}}(p) \\
    &\mathcal{L}_{\text{UAL}}(p) = (1 - |2p - 1|^2)
\end{aligned}
\label{eq:stage2_loss}
\end{equation}
where $\mathcal{L}_{\text{BCE}}$ is the binary cross-entropy loss, the uncertainty-aware loss $\mathcal{L}_{\text{UAL}}$ penalizes uncertain predictions, and $\alpha(t) = \cos(\pi t / 2)$ is a cosine-annealed coefficient with $t \in [0,1]$ denoting the training progress. This design encourages the model to produce confident predictions while leveraging the high-quality pseudo-labels from the first stage.

\input{table01_modify}

%% file: table01_modify.tex
\begin{table*}[htbp]
\centering
\renewcommand{\arraystretch}{1.15}
\caption{Comprehensive evaluation of FCL-COD against competing methods across four benchmarks. \textbf{Sup.}: F = fully supervised, S = scribble, B = bounding box, P = point, -- = zero-shot/no task-specific training.}
\resizebox{1\textwidth}{!}
{
\begin{tabular}{l|c|cccc|cccc|cccc|cccc}
\hline
\multirow{2}{*}{\textbf{Methods}} &
  \multirow{2}{*}{\textbf{Sup.}} &
  \multicolumn{4}{c|}{\textbf{CAMO}} &
  \multicolumn{4}{c|}{\textbf{COD10K}} &
  \multicolumn{4}{c|}{\textbf{NC4K}} &
  \multicolumn{4}{c}{\textbf{CHAMELEON}} \\ \cline{3-18}
 &
   &
  \textbf{MAE↓} &
  \textbf{S$_m$↑} &
  \textbf{E$_m$↑} &
  \textbf{F$_{\beta}^w$↑} &
  \textbf{MAE↓} &
  \textbf{S$_m$↑} &
  \textbf{E$_m$↑} &
  \textbf{F$_{\beta}^w$↑} &
  \textbf{MAE↓} &
  \textbf{S$_m$↑} &
  \textbf{E$_m$↑} &
  \textbf{F$_{\beta}^w$↑} &
  \textbf{MAE↓} &
  \textbf{S$_m$↑} &
  \textbf{E$_m$↑} &
  \textbf{F$_{\beta}^w$↑} \\ 
\hline
\textbf{UGTR\cite{yang2021uncertainty}} &
   &
  0.086 &
  0.784 &
  0.822 &
  0.684 &
  0.036 &
  0.817 &
  0.852 &
  0.666 &
  0.052 &
  0.839 &
  0.874 &
  0.747 &
  0.030 &
  0.891 &
  0.955 &
  0.833 \\
\textbf{ZoomNet\cite{pang2022zoom}} &
   &
  0.066 &
  0.820 &
  0.892 &
  0.752 &
  0.029 &
  0.838 &
  0.911 &
  0.729 &
  0.043 &
  0.853 &
  0.896 &
  0.784 &
  0.023 &
  0.902 &
  0.958 &
  0.845 \\
\textbf{SAM-Adapter\cite{chen2023sam}} &
   &
  0.070 &
  0.847 &
  0.873 &
  0.765 &
  0.025 &
  0.883 &
  0.918 &
  0.801 &
  - &
  - &
  - &
  - &
  0.033 &
  0.896 &
  0.919 &
  0.824 \\
\textbf{CamoFormer-R\cite{yin2024camoformer}} &
   &
  0.067 &
  0.817 &
  0.885 &
  0.752 &
  0.029 &
  0.838 &
  0.930 &
  0.724 &
  0.042 &
  0.855 &
  0.914 &
  0.788 &
  0.025 &
  0.898 &
  0.956 &
  0.847 \\
\textbf{FEDER\cite{he2023camouflaged}} &
   &
  0.071 &
  0.802 &
  0.873 &
  0.738 &
  0.032 &
  0.822 &
  0.905 &
  0.716 &
  0.044 &
  0.847 &
  0.915 &
  0.789 &
  0.030 &
  0.887 &
  0.954 &
  0.835 \\
\textbf{CamoFormer-P\cite{yin2024camoformer}} &
   &
  0.046 &
  0.872 &
  0.938 &
  0.831 &
  0.023 &
  0.869 &
  0.939 &
  0.786 &
  0.030 &
  0.892 &
  0.946 &
  0.847 &
  0.022 &
  0.910 &
  0.966 &
  0.865 \\
\textbf{MSCAF-Net\cite{liu2023mscaf}} &
   &
  0.046 &
  0.873 &
  0.937 &
  0.828 &
  0.024 &
  0.865 &
  0.936 &
  0.775 &
  0.032 &
  0.887 &
  0.942 &
  0.838 &
  0.022 &
  0.912 &
  0.970 &
  0.865 \\
\textbf{HitNet\cite{hu2023high}} &
   &
  0.055 &
  0.849 &
  0.910 &
  0.809 &
  0.023 &
  0.871 &
  0.938 &
  0.806 &
  0.037 &
  0.875 &
  0.929 &
  0.834 &
  0.019 &
  0.921 &
  0.972 &
  0.897 \\
\textbf{FSPNet\cite{huang2023feature}} &
   &
  0.050 &
  0.856 &
  0.928 &
  0.799 &
  0.026 &
  0.851 &
  0.930 &
  0.735 &
  0.035 &
  0.878 &
  0.937 &
  0.816 &
  0.023 &
  0.908 &
  0.965 &
  0.851 \\
\textbf{SARNet\cite{xing2023go}} &
  \multirow{-10}{*}{F} &
  0.046 &
  0.874 &
  0.935 &
  0.844 &
  0.021 &
  0.885 &
  0.947 &
  0.820 &
  0.032 &
  0.889 &
  0.940 &
  0.851 &
  0.017 &
  0.933 &
  0.978 &
  0.909 \\ 
\hline
\textbf{SAM\cite{kirillov2023segment} (SAM-H)} &
  - &
  0.132 &
  0.684 &
  0.687 &
  0.606 &
  0.050 &
  0.783 &
  0.798 &
  0.701 &
  0.078 &
  0.767 &
  0.776 &
  0.696 &
  0.081 &
  0.727 &
  0.734 &
  0.639 \\ 
\hline
\textbf{SCSOD\cite{yu2021structure}} &
   &
  0.102 &
  0.713 &
  0.795 &
  0.618 &
  0.055 &
  0.710 &
  0.805 &
  0.546 &
  - &
  - &
  - &
  - &
  0.053 &
  0.792 &
  0.881 &
  0.714 \\
\textbf{CRNet\cite{he2023weakly}} &
   &
  0.092 &
  0.735 &
  0.815 &
  0.641 &
  0.049 &
  0.733 &
  0.832 &
  0.576 &
  0.063 &
  0.775 &
  0.855 &
  0.688 &
  0.046 &
  0.818 &
  0.897 &
  0.791 \\
\textbf{SAM-S\cite{kirillov2023segment} (SAM-H)} &
   &
  0.105 &
  0.731 &
  0.774 &
  - &
  0.046 &
  0.772 &
  0.828 &
  - &
  0.071 &
  0.763 &
  0.832 &
  - &
  0.076 &
  0.650 &
  0.820 &
  0.729 \\
\textbf{WS-SAM\cite{he2024weakly} (SAM-H)} &
  \multirow{-4}{*}{S} &
  0.092 &
  0.759 &
  0.818 &
  - &
  0.038 &
  0.803 &
  0.878 &
  - &
  0.052 &
  0.829 &
  0.886 &
  - &
  0.046 &
  0.824 &
  0.897 &
  0.777 \\ 
\hline
\textbf{SAM-P\cite{kirillov2023segment} (SAM-H)} &
   &
  0.123 &
  0.677 &
  0.693 &
  - &
  0.069 &
  0.765 &
  0.796 &
  - &
  0.082 &
  0.776 &
  0.786 &
  - &
  0.101 &
  0.697 &
  0.745 &
  0.696 \\
\textbf{WS-SAM\cite{he2024weakly} (SAM-H)} &
  \multirow{-2}{*}{P} &
  0.102 &
  0.718 &
  0.757 &
  - &
  0.039 &
  0.790 &
  0.856 &
  - &
  0.057 &
  0.813 &
  0.859 &
  - &
  0.056 &
  0.805 &
  0.868 &
  0.767 \\ 
\hline
\textbf{SAM-COD\cite{chen2024sam}(SAM-H)} &
   &
  0.062 &
  0.837 &
  0.901 &
  0.786 &
  0.028 &
  0.842 &
  0.914 &
  0.745 &
  0.037 &
  0.867 &
  0.923 &
  0.813 &
  - &
  - &
  - &
  - \\
\textbf{FCL-COD(SAM-B)} &
   &
  0.060 &
  0.841 &
  0.899 &
  0.795 &
  0.027 &
  0.859 &
  0.924 &
  0.774 &
  0.041 &
  0.867 &
  0.919 &
  0.817 &
  0.039 &
  0.866 &
  0.928 &
  0.799 \\
\textbf{FCL-COD(SAM-L)} &
   &
  0.054 & 0.856 & 0.910 & 0.818 & 
  \textbf{0.022} & \textbf{0.881} & \textbf{0.938} & \textbf{0.812} & 
  0.034 & \textbf{0.886} & \textbf{0.930} & \textbf{0.847} & 
  \textbf{0.026} & \textbf{0.901} & \textbf{0.954} & \textbf{0.856} \\
\textbf{FCL-COD(SAM-H)} &
  \multirow{-4}{*}{B} &
  \textbf{0.050} & \textbf{0.862} & \textbf{0.915} & \textbf{0.824} & 
  \textbf{0.022} & 0.878 & 0.934 & 0.808 & 
  \textbf{0.033} & 0.885 & 0.928 & 0.846 & 
  0.038 & 0.882 & 0.932 & 0.842 \\ 
\hline
\end{tabular}
}
\label{tab:1}
\end{table*}

%% file: sec/4_experiment.tex
\section{Experiment}
\label{experiment}
\subsection{Experimental Setup}
\noindent\textbf{Dataset}: We conducted experiments on four widely used COD benchmarks: CAMO~\cite{le2019anabranch}, CHAMELEON~\cite{skurowski2018animal}, COD10K~\cite{fan2020camouflaged}, and NC4K~\cite{lv2021simultaneously}. CAMO includes 1,250 camouflaged and 1,250 non-camouflaged images, while CHAMELEON provides 76 finely annotated samples. COD10K is a large-scale dataset with 5,066 camouflaged, 3,000 background, and 1,934 non-camouflaged images, offering high diversity. NC4K contains 4,121 images collected from multiple online sources. Following~\cite{mei2021camouflaged}, we use only camouflaged images. Specifically, 3,040 images from COD10K and 1,000 from CAMO form the training set, while the remaining camouflaged images from all four datasets serve as the test set.

\noindent\textbf{Implementation Details}: Our method is implemented in PyTorch on two NVIDIA H20 GPUs. Stage 1 adapts SAM via FoRA for pseudo-label generation; Stage 2 uses PVT-B4 as the lightweight encoder-decoder backbone. Training uses SGD with momentum 0.9, weight decay $5\times10^{-4}$, and cosine annealing from $1\times10^{-3}$ over 60 epochs with batch size 8. Bounding-box prompts are derived from ground-truth mask bounding boxes; no pixel-level annotations are used. At inference, Stage 2 operates without prompt input.

\begin{figure}[t]
    \centering
    \includegraphics[width=1.0\linewidth]{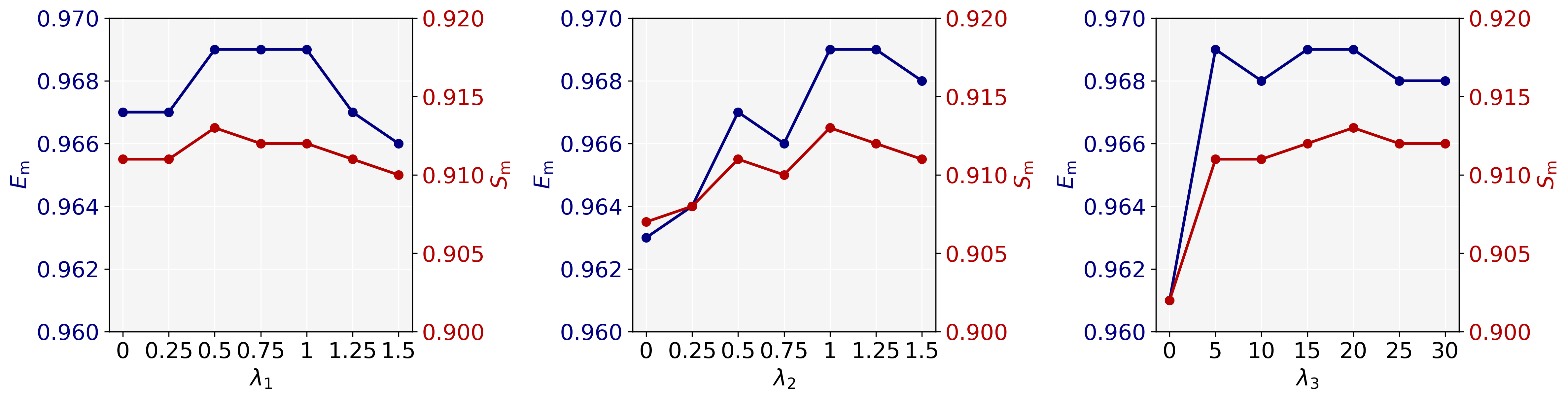}
    \caption{Hyperparameter analysis of loss-function weights.}
    \label{fig:hyper_analysis}
    \end{figure}

\subsection{Comparison with State-of-the-art Methods}
 \noindent\textbf{Quantitative Comparison}: As demonstrated in \cref{tab:1}, our method delivers substantial improvements over existing approaches. When compared to the state-of-the-art weakly-supervised COD method, SAM-COD, our method consistently outperforms across all evaluation metrics. Specifically, on the CAMO dataset, our approach results in a reduction of the MAE by 0.012, alongside improvements in $S_m$, $E_m$, and $F^w_{\beta}$ by 0.025, 0.014, and 0.038, respectively. These performance gains are not only observed in CAMO but also extend consistently across the remaining three datasets. Furthermore, when contrasted with fully-supervised methods such as ZoomNet and CamoFormer, our method achieves noticeable performance advancements across all four datasets.
 
\begin{figure*}[!h]
\centering
\includegraphics[width=0.8\linewidth]{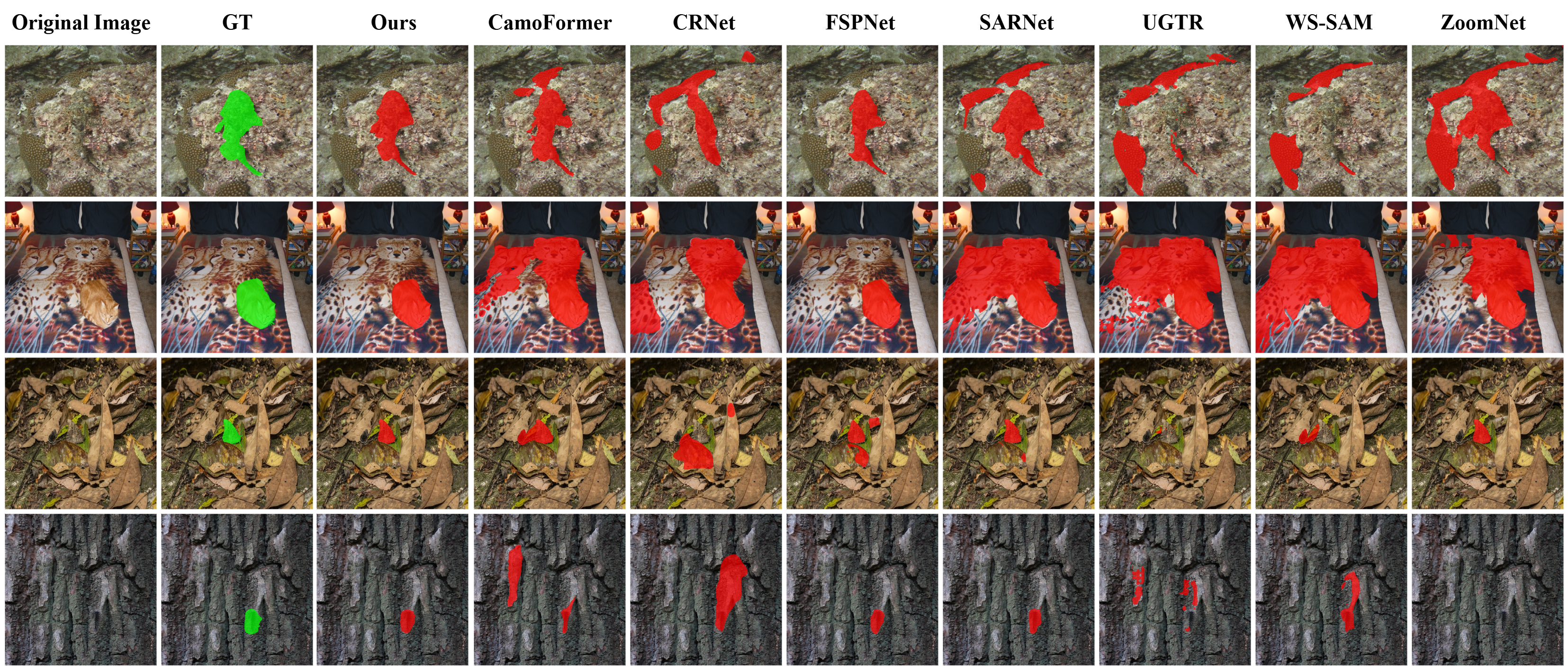}

\caption{Qualitative Comparison between FCL-COD and Competing Methods}
\label{fig:visual_result}
\end{figure*}
\noindent\textbf{Qualitative Comparison}: As illustrated in \cref{fig:visual_result}, the predicted maps produced by our method demonstrate clearer, more coherent object regions with more defined contours. These results surpass the performance of the state-of-the-art weakly-supervised COD method, SAM-COD, and the fully-supervised COD method, ZoomNet. Our approach effectively addresses limitations observed in prior methods, including the presence of non-camouflaged target responses, extreme or partial responses, and rough boundary delineations.

\subsection{Ablation Study}

\input{table02_modify}
\noindent\textbf{Ablation experiment of components.}We conducted a progressive ablation study to evaluate each component’s contribution, as shown in \cref{tab:IC}. The pseudo-label quality on \noindent\textbf{COD-train} steadily improves with the integration of \noindent\textbf{FoRA} and \noindent\textbf{GCL}, boosting the $E_m$ from 0.959 to 0.969. This enhancement directly benefits the final lightweight model, where the full system with \noindent\textbf{MSFA} achieves the best results ($E_m$=0.938 on \noindent\textbf{COD10K}, 0.954 on \noindent\textbf{CHAMELEON}), validating the synergy of all modules.

\input{table05_modify}

\noindent\textbf{Hyperparameter Analysis.}
We analyze the loss-weight parameters in \cref{fig:hyper_analysis}. Optimal values are $\lambda_1$=0.50, $\lambda_2$=1.00, and $\lambda_3$=20.

\begin{figure}
    \centering
    \includegraphics[width=0.95\linewidth]{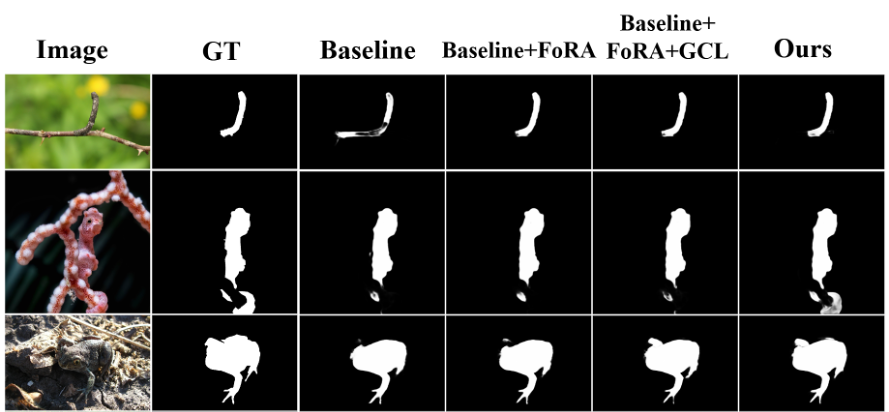}
    \caption{Qualitative ablation study of each component.}
    \label{fig:ablation_vis}
\end{figure}

\noindent\textbf{Ablation Study of FoRA.}
As shown in \cref{tab:table3}, the baseline attains an $E_m$ of 0.965. Incorporating either the frequency modulation stage $\mathcal{S}{\text{fre}}$ or the spatial enhancement stage $\mathcal{S}{\text{spa}}$ improves performance to 0.967 and 0.966, respectively. The complete FoRA, combining both stages, achieves the highest $E_m$ of 0.969, validating the complementary contributions of spatial and frequency cues to fine-grained segmentation.

\noindent\textbf{Ablation Study of GCL.}
To assess Gradient-aware Contrastive Learning (GCL), we conduct a controlled ablation in \cref{tab:table3}. The baseline (\textbf{noCL}) yields an $E_m$ of 0.963, while standard \textbf{CL} increases it to 0.968. Our \textbf{GCL} leverages gradient cues to highlight hard samples, further improving performance to 0.969.

\noindent\textbf{Ablation Study of MSFA.}
\Cref{tab:table2} shows that the \textbf{Base} model reaches $E_m$ scores of 0.926 (COD10K) and 0.944 (CHAMELEON). Introducing the single-scale Tri-Channel Attention $\mathcal{T}$ provides moderate gains, while adding the spatial branch $\mathcal{M}{\text{spa}}$ and frequency branch $\mathcal{M}{\text{fre}}$ yields further improvements. The full multi-scale design delivers the largest boost, with MSFA (\textbf{Ours}) achieving 0.938 and 0.954, confirming that multi-scale fusion drives the majority of the performance gains, complemented by dual-branch attention.


\begin{figure}
    \centering
    \includegraphics[width=0.95\linewidth]{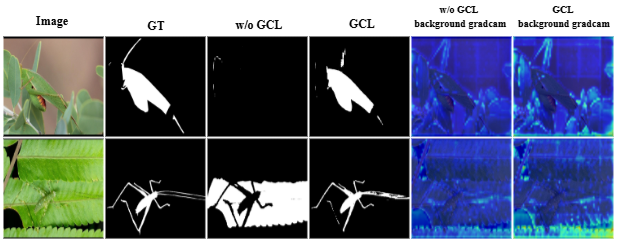}
    \caption{Qualitative analysis of GCL.}
    \label{fig:Grad_CAM}
\end{figure}

\begin{figure}
    \centering
    \includegraphics[width=0.95\linewidth]{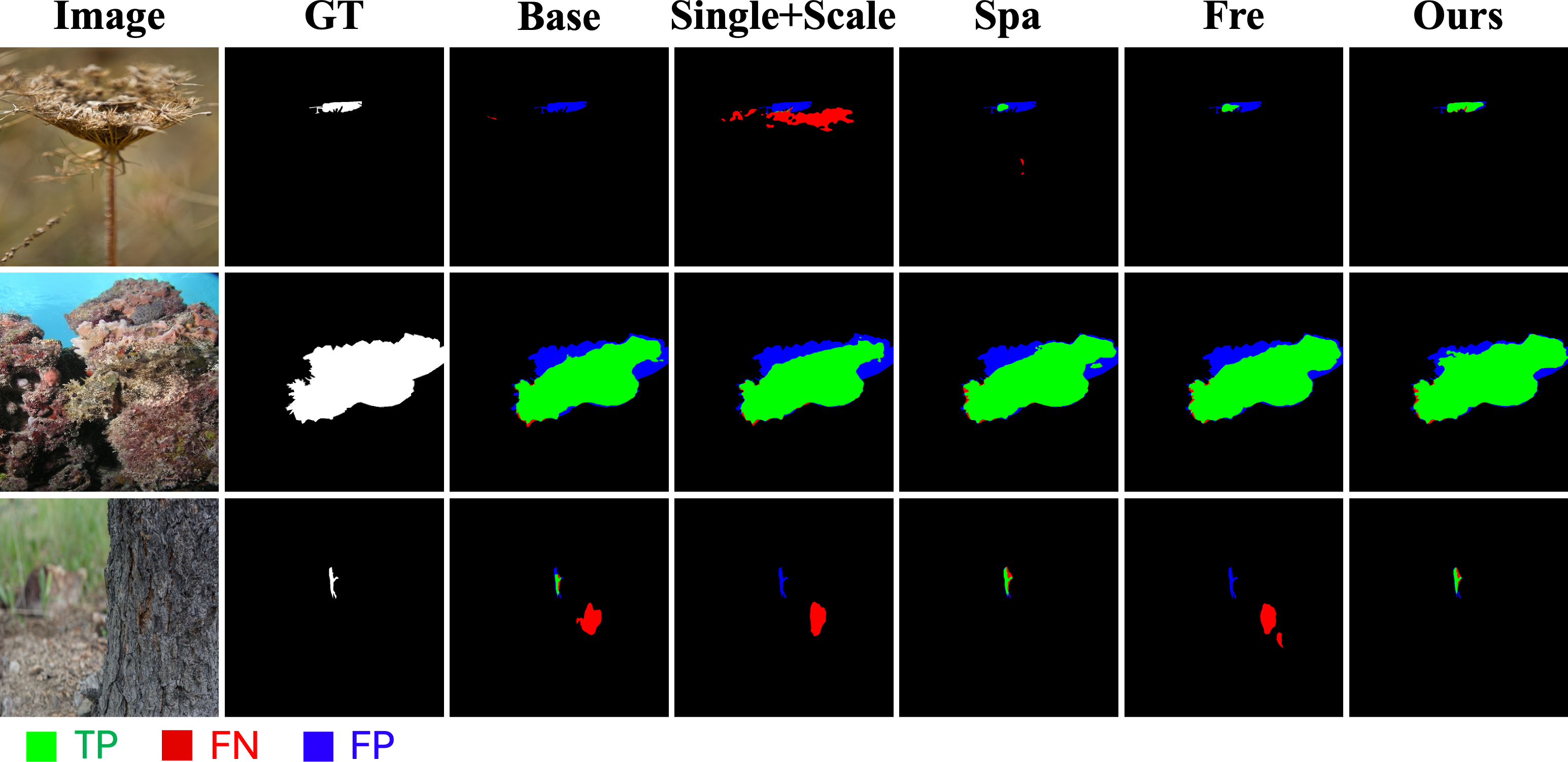}
    \caption{Qualitative analysis of MSFA.}
    \label{fig:msfa_vis}
\end{figure}

\begin{figure}
    \centering
    \includegraphics[width=0.95\linewidth]{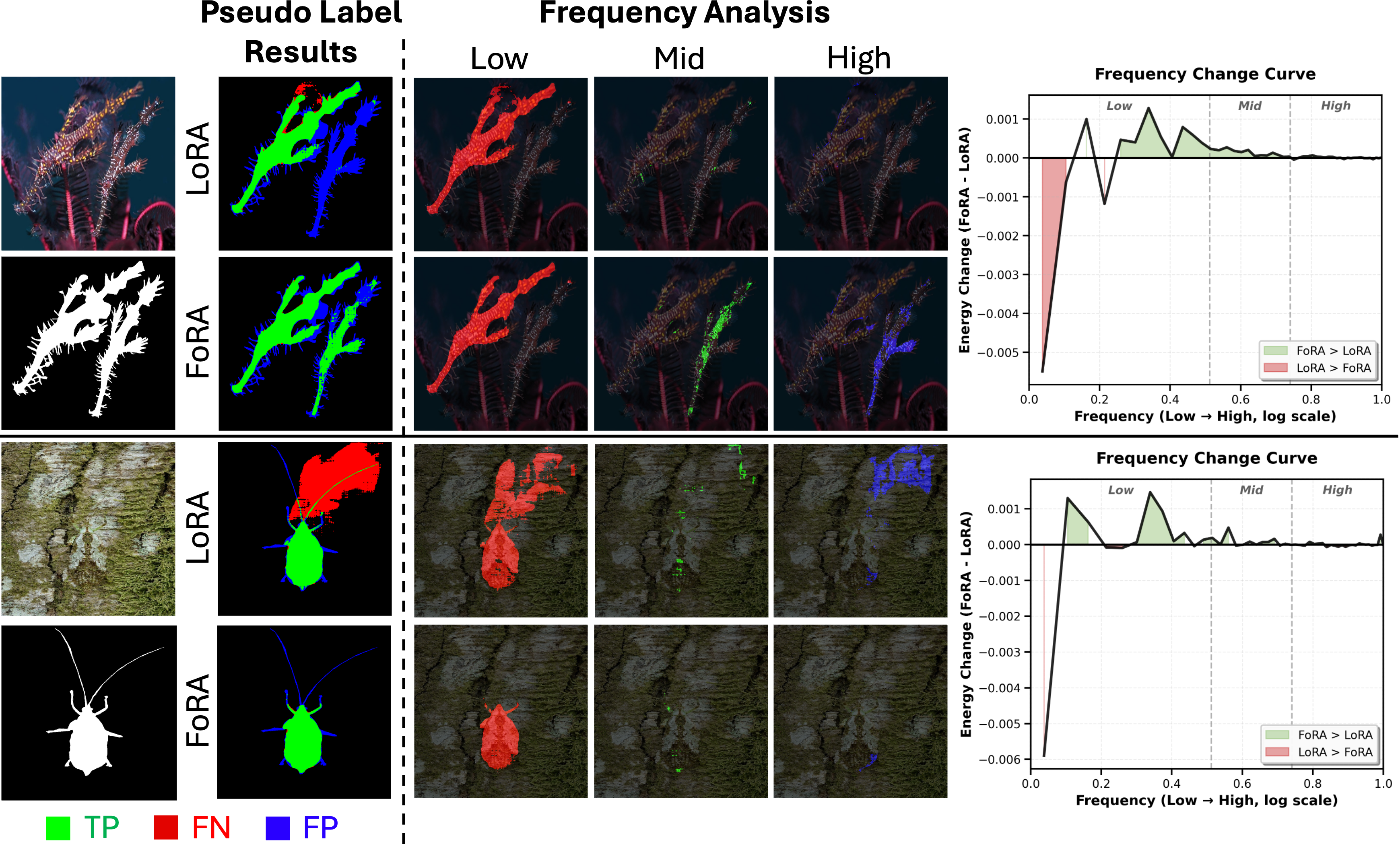}
    \caption{Qualitative analysis of frequency awareness achieved by FoRA.}
    \label{fig:fora_exp}
\end{figure}


\noindent\textbf{Qualitative Analysis for Ablation Study.}
\Cref{fig:ablation_vis} illustrates the progressive gains from each component. The \textbf{Baseline} produces coarse and incomplete structures. Incorporating \textbf{FoRA} improves core-region localization, yielding more complete predictions. Adding \textbf{GCL} further enhances object–background separation and contour refinement. With MSFA integrated, the \textbf{Full Model} achieves accurate segmentation with well-defined boundaries through multi-scale feature fusion. These results align with quantitative trends, validating each module's contribution.

\noindent\textbf{Effectiveness of the GCL module with Grad-CAM.}
\label{sec:gcl_effectiveness}
\cref{fig:Grad_CAM} demonstrates the impact of GCL. Grad-CAM results show that \noindent\textbf{GCL} effectively guides attention to background regions that resemble the foreground—critical hard negatives for contrastive learning—whereas \noindent\textbf{noGCL} suffers from dispersed focus. This targeted hard-sample awareness leads to more accurate and robust segmentation.

\input{table07_con}

\noindent\textbf{Qualitative Analysis of MSFA.}
As shown in \cref{fig:msfa_vis}, the Base model suffers from localized responses (row 1 \& 3) and coarse boundaries. Progressively integrating spatial and frequency branches reduces false positives, while the full MSFA achieves complete object coverage with refined boundaries through multi-scale spatial–frequency fusion.

\noindent\textbf{Qualitative Analysis of FoRA.}
\cref{fig:fora_exp} demonstrates that vanilla LoRA produces extreme responses (excessive false negatives in rows 1-3) and non-camouflaged responses. In contrast, FoRA substantially mitigates these issues by suppressing low-frequency texture interference while preserving discriminative mid–high frequency details, as evidenced by the feature frequency amplitude spectra, which show notable energy redistribution toward mid-to-high frequency components.

\noindent\textbf{Generalization to salient object detection.}
\cref{tab:my-table-2} further shows that our framework generalizes well to salient object detection (SOD). Enhanced frequency perception and contrastive learning strengthen SAM’s adaptability to complex scenes, highlighting the versatility of the proposed weakly supervised paradigm.

%% file: table02_modify.tex
\begin{table*}[ht]
\centering
\caption{Ablation study of Individual Components}
\label{tab:IC}
\resizebox{1\textwidth}{!}{
\begin{tabular}{ccc|cccccc|cccccc|cccccc}
\hline
\multirow{2}{*}{\textbf{FoRA}} &
  \multirow{2}{*}{\textbf{GCL}} &
  \multirow{2}{*}{\textbf{MSFA}} &
  \multicolumn{6}{c|}{\textbf{COD-Train}} &
  \multicolumn{6}{c|}{\textbf{CHAMELLEON}}
  &
  \multicolumn{6}{c}{\textbf{COD10K}}\\ \cline{4-21}  
 &
   &
   &
  \textbf{MAE↓} &
  \textbf{S$_m$↑} &
  \textbf{E$_m$↑} &
  \textbf{F$_{\beta}^w$↑} &
  \textbf{mIOU↑} &
  \textbf{mF1↑} &
   \textbf{MAE↓} &
  \textbf{S$_m$↑} &
  \textbf{E$_m$↑} &
  \textbf{F$_{\beta}^w$↑} &
  \textbf{mIOU↑} &
  \textbf{mF1↑} &
  \textbf{MAE↓} &
  \textbf{S$_m$↑} &
  \textbf{E$_m$↑} &
  \textbf{F$_{\beta}^w$↑} &
  \textbf{mIOU↑} &
  \textbf{mF1↑} \\ \hline
\text{\ding{55}} & \text{\ding{55}} & \text{\ding{55}} & 0.017 & 0.900   & 0.959 & 0.865 & 0.799 & 0.874& 0.041 & 0.864 & 0.927 & 0.801 & 0.745 & 0.826 &  0.025 & 0.856 & 0.919 & 0.766 & 0.702 & 0.794 \\
\text{\ding{51}} & \text{\ding{55}} & \text{\ding{55}} & 0.015 & 0.907 & 0.963 & 0.878 & 0.812 & 0.884& 0.041 & 0.868 & 0.928 & 0.809 & 0.753 & 0.833 &0.024&0.860&0.923&0.775&0.709&0.800\\
\text{\ding{51}} & \text{\ding{51}} & \text{\ding{55}} & \textbf{0.013} & \textbf{0.913} & \textbf{0.969} & \textbf{0.887} & \textbf{0.820} & \textbf{0.892} & 0.029 & 0.888 & 0.947 & 0.839 & 0.778 & 0.857& 0.024 & 0.863 & 0.926 & 0.782 & 0.716 & 0.807    \\
\text{\ding{51}} & \text{\ding{51}} & \text{\ding{51}} &  - &- &- &- &- &-& \textbf{0.026} & \textbf{0.901} & \textbf{0.954} & \textbf{0.856} & \textbf{0.801} & \textbf{0.873} & \textbf{0.022} & \textbf{0.881} & \textbf{0.938} & \textbf{0.812} & \textbf{0.750} & \textbf{0.836}\\ \hline
\end{tabular}
}

\end{table*}

%% file: table05_modify.tex
\begin{table*}[h]
\centering
\begin{minipage}[b]{0.36\textwidth}
    \caption{Ablation study on FoRA and GCL}
    \label{tab:table3}
    \centering
    \resizebox{\textwidth}{!}{
    \begin{tabular}{cc|cccccc}
\hline
\multicolumn{2}{c|}{\textbf{FoRA}} & \multicolumn{6}{c}{\textbf{COD-Train}} \\ \hline
$\mathcal{S}_{\text{spa}}$ & $\mathcal{S}_{\text{fre}}$ & \textbf{MAE↓} & \textbf{S$_m$↑} & \textbf{E$_m$↑} & \textbf{F$_{\beta}^w$↑} & \textbf{mIOU↑} & \textbf{mF1↑} \\ \hline
\text{\ding{55}} & \text{\ding{55}} & 0.014 & 0.91  & 0.965 & 0.881 & 0.814 & 0.885 \\
\text{\ding{55}} & \text{\ding{51}} & 0.014 & 0.91  & 0.967 & 0.883 & 0.816 & 0.888 \\
\text{\ding{51}} & \text{\ding{55}} & 0.014 & 0.91  & 0.966 & 0.882 & 0.817 & 0.888 \\
\text{\ding{51}} & \text{\ding{51}} & \textbf{0.013} & \textbf{0.913} & \textbf{0.969} & \textbf{0.887} & \textbf{0.82}  & \textbf{0.892} \\ 
\hline
\multicolumn{2}{c|}{\textbf{GCL}} & \multicolumn{6}{c}{\textbf{COD-Train}} \\ \hline
\textbf{CL} & \textbf{Grad-aware} & \textbf{MAE↓} & \textbf{S$_m$↑} & \textbf{E$_m$↑} & \textbf{F$_{\beta}^w$↑} & \textbf{mIOU↑} & \textbf{mF1↑} \\ \hline
\text{\ding{55}} & \text{\ding{55}} & 0.015 & 0.907 & 0.963 & 0.878 & 0.812 & 0.884 \\
\text{\ding{51}} & \text{\ding{55}} & 0.014 & 0.912 & 0.968 & 0.883 & 0.817 & 0.89  \\
\text{\ding{51}} & \text{\ding{51}} & \textbf{0.013} & \textbf{0.913} & \textbf{0.969} & \textbf{0.887} & \textbf{0.82}  & \textbf{0.892} \\ \hline
\end{tabular}
    }

\end{minipage} \hfill
\begin{minipage}[b]{0.62\textwidth}
    \centering
        \caption{Ablation Study on MSFA}
    \label{tab:table2}
    \resizebox{\textwidth}{!}{
    \begin{tabular}{ccc|cccccc|cccccc}
\hline
\multicolumn{3}{c|}{\textbf{MSFA}} & \multicolumn{6}{c|}{\textbf{COD10K}} & \multicolumn{6}{c}{\textbf{CHAMELLEON}} \\ \hline
$\mathcal{T}$ & $\mathcal{M}_{\text{spa}}$ & $\mathcal{M}_{\text{fre}}$ &
  \textbf{MAE↓} &
  \textbf{S$_m$↑} &
  \textbf{E$_m$↑} &
  \textbf{F$_{\beta}^w$↑} &
  \textbf{mIOU↑} &
  \textbf{mF1↑} &
  \textbf{MAE↓} &
  \textbf{S$_m$↑} &
  \textbf{E$_m$↑} &
  \textbf{F$_{\beta}^w$↑} &
  \textbf{mIOU↑} &
  \textbf{mF1↑} \\ \hline
\text{\ding{55}} & \text{\ding{55}} & \text{\ding{55}} & 0.024 & 0.863 & 0.926 & 0.782 & 0.716 & 0.807 & 0.03  & 0.886 & 0.944 & 0.835 & 0.773 & 0.85  \\
\text{\ding{51}} & \text{\ding{55}} & \text{\ding{55}} & 0.024 & 0.865 & 0.927 & 0.785 & 0.718 & 0.809 & 0.03  & 0.887 & 0.945 & 0.836 & 0.774 & 0.852 \\
\text{\ding{51}} & \text{\ding{51}} & \text{\ding{55}} & \textbf{0.022} & 0.88  & 0.936 & 0.811 & 0.749 & 0.834 & 0.027 & 0.898 & 0.947 & 0.852 & 0.796 & 0.869 \\
\text{\ding{51}} & \text{\ding{55}} & \text{\ding{51}} & \textbf{0.022} & 0.877 & 0.934 & 0.807 & 0.743 & 0.829 & 0.027 & 0.895 & 0.95  & 0.85  & 0.792 & 0.865 \\
\text{\ding{51}} & \text{\ding{51}} & \text{\ding{51}} & \textbf{0.022} & \textbf{0.881} & \textbf{0.938} & \textbf{0.812} & \textbf{0.75} & \textbf{0.836} & \textbf{0.026} & \textbf{0.901} & \textbf{0.954} & \textbf{0.856} & \textbf{0.801} & \textbf{0.873} \\ \hline
\end{tabular}
    }

\end{minipage}
\end{table*}

%% file: table07_con.tex
\begin{table}[t]
\centering
\caption{Extended Experiments of FCL-COD on Weakly-Supervised Salient Object Detection}
\label{tab:my-table-2}
\resizebox{0.45\textwidth}{!}{%
\begin{tabular}{c|c|cc|cc|cc|cc}
\hline
                      &                          & \multicolumn{2}{c|}{\textbf{ECSSD}} & \multicolumn{2}{c|}{\textbf{DUT-O}} & \multicolumn{2}{c|}{\textbf{HKU-IS}} & \multicolumn{2}{c}{\textbf{DUTS-TE}} \\ \cline{3-10} 
\multirow{-2}{*}{}    & \multirow{-2}{*}{\textbf{Label}} & \textbf{MAE↓} & \textbf{S$_m$↑} & \textbf{MAE↓} & \textbf{S$_m$↑} & \textbf{MAE↓} & \textbf{S$_m$↑} & \textbf{MAE↓} & \textbf{S$_m$↑} \\ \hline
\textbf{AFNet\cite{liu2020afnet}}        & F                        & 0.042         & 0.913         & 0.057         & 0.826         & 0.036         & 0.905         & 0.046         & 0.867         \\
\textbf{BASNet\cite{qin2019basnet}}       & F                        & 0.037         & 0.916         & 0.057         & 0.836         & 0.032         & 0.909         & 0.048         & 0.866         \\
\textbf{GateNet}\cite{zhao2024towards}      & F                        & 0.040         & 0.920         & 0.055         & 0.838         & 0.033         & 0.915         & 0.040         & 0.885         \\
\textbf{ICON-R\cite{zhuge2022salient}}       & F                        & 0.032         & 0.928         & 0.057         & 0.845         & 0.029         & 0.920         & 0.037         & 0.890         \\
\textbf{MENet\cite{wang2023pixels}}        & F                        & 0.031         & 0.928         & \underline{0.045} & 0.850         & \underline{0.023} & 0.927         & \textbf{0.028} & 0.905         \\
\textbf{VST-S++ \cite{liu2024vst++}}     & F                        & \underline{0.027} & \underline{0.939} & 0.050         & \underline{0.859} & 0.025         & 0.932         & \underline{0.029} & \textbf{0.909} \\
\textbf{SCSOD\cite{yu2021structure}}        & S                        & 0.049         & 0.881         & 0.060         & 0.811         & 0.038         & 0.882         & 0.049         & 0.853         \\
\textbf{PSOD\cite{gao2022weakly}}         & P                        & 0.036         & 0.913         & 0.064         & 0.824         & 0.033         & 0.901         & 0.045         & 0.853         \\
\textbf{SAM-COD\cite{chen2024sam}}      & B                        & 0.031         & 0.929         & 0.051         & 0.844         & \underline{0.023} & \textbf{0.952} & 0.033         & 0.899         \\
\textbf{FCL-COD}      & B                        & \textbf{0.023} & \textbf{0.945} & \textbf{0.043} & \textbf{0.872} & \textbf{0.020} & \underline{0.942} & 0.030         & \underline{0.907} \\ \hline
\end{tabular}%
}
\end{table}

%% file: sec/5_conclusion.tex
\section{Conclusion}
\label{conclusion}

We propose \textbf{FCL-COD}, a frequency-aware and contrastive learning-based framework that addresses key limitations in weakly supervised camouflaged object detection: non-camouflaged responses, localized and extreme responses, and coarse boundaries. Through frequency-aware low-rank adaptation (FoRA), gradient-aware contrastive learning (GCL), and multi-scale frequency-aware attention (MSFA), FCL-COD effectively adapts SAM to camouflage scenarios with sparse annotations. Extensive experiments demonstrate that our method surpasses existing weakly supervised approaches and rivals fully supervised methods, highlighting the potential of frequency-domain modeling and contrastive learning for challenging perception tasks.